\documentclass[letterpaper, 10 pt, journal, twoside]{ieeetran}

\IEEEoverridecommandlockouts

\usepackage{epsfig}
\usepackage{graphicx,epstopdf,wrapfig}
\usepackage{comment}
\usepackage{algorithm}
\usepackage{algorithmic}
\usepackage{amsmath}
\usepackage{amsfonts}
\usepackage{amssymb}
\usepackage{gensymb}
\usepackage{mdwlist,url}
\usepackage{xcolor}
\usepackage{amsthm}
\usepackage{wrapfig}
\usepackage{tikz}
\usepackage{nicefrac}
\usepackage{bm}
\usepackage{cite} 
\usepackage{textcomp}
\usepackage[utf8]{inputenc}
\usetikzlibrary{automata,positioning,shapes,arrows}
\usepackage{pifont}
\newcommand{\cmark}{\textcolor{green}{\ding{51}}}%
\newcommand{\xmark}{\textcolor{red}{\ding{55}}}%

\makeatletter
\let\MYcaption\@makecaption
\makeatother

\usepackage[font=footnotesize]{subcaption}

\makeatletter
\let\@makecaption\MYcaption
\makeatother

\newcommand{\x}{\mathbf{x}}
\newcommand{\xdot}{\dot{\x}}
\newcommand{\y}{\mathbf{y}}
\renewcommand{\u}{\mathbf{u}}
\newcommand{\J}{\mathbf{J}}
\newcommand{\M}{\mathbf{M}}
\newcommand{\V}{\mathcal{V}}
\newcommand{\Vdot}{\dot{\V}}
\newcommand{\K}{\mathbf{K}}
\newcommand{\q}{\mathbf{q}}
\newcommand{\qdot}{\dot{\q}}
\newcommand{\qddot}{\ddot{\q}}
\newcommand{\btau}{\boldsymbol{\tau}}
\newcommand{\R}{\mathbb{R}}
\newcommand{\LL}{\boldsymbol{\Lambda}}
\newcommand{\z}{\mathbf{z}}

\newtheorem{definition}{Definition}
\newtheorem{theorem}{Theorem}
\newtheorem{corollary}{Corollary}
\newtheorem{lemma}{Lemma}
\newtheorem{remark}{Remark}

\title{Approximate Simulation for Template-Based Whole-Body Control }

\author{Vince Kurtz$^{1}$, Patrick M. Wensing$^{2}$, and Hai Lin$^{1}$%
\thanks{Manuscript received: July 17, 2020; Revised November 11, 2020; Accepted December 9, 2020.}
\thanks{This paper was recommended for publication by Editor Abderrahmane Kheddar upon evaluation of the Associate Editor and Reviewers’ comments.}
\thanks{This work was supported by NSF Grants IIS-1724070, CNS-1830335, IIS-2007949, CMMI-1835186.}%
\thanks{$^{1} $V. Kurtz and H. Lin are with the Department of Electrical Engineering, University of Notre Dame, Notre Dame, IN, 46556 USA. \texttt{\{vkurtz,hlin1\}@nd.edu}.}%
\thanks{$^{2} $P. M. Wensing is with the Department of Aerospace and Mechanical Engineering, University of Notre Dame, Notre Dame, IN, 46556 USA. \texttt{pwensing@nd.edu}.}
\thanks{Digital Object Identifier (DOI): see top of this page.}
}

\begin{document}

\maketitle

\begin{abstract}
    Reduced-order template models are widely used to control high degree-of-freedom legged robots, but existing methods for template-based whole-body control rely heavily on heuristics and often suffer from robustness issues. In this letter, we propose a template-based whole-body control method grounded in the formal framework of approximate simulation. Our central contribution is to demonstrate how the Hamiltonian structure of rigid-body dynamics can be exploited to establish approximate simulation for a high-dimensional nonlinear system. The resulting controller is passive, more robust to push disturbances, uneven terrain, and modeling errors than standard QP-based methods, and naturally enables high center of mass walking. Our theoretical results are supported by simulation experiments with a 30 degree-of-freedom Valkyrie humanoid model. 
\end{abstract}


\section{Introduction}\label{sec:intro}

\IEEEPARstart{R}{educed}-order ``template'' models are foundational to controlling high degree-of-freedom legged robots {\cite{wieber2016modeling,kuindersma2016optimization,wensing2017template,liu2019leveraging}}. Such models, like the Linear Inverted Pendulum (LIP), abstract away complicating details of a robot's dynamics, enabling focus on key aspects of balancing and locomotion, such as regulating the center of mass (CoM) and center of pressure (CoP) \cite{kajita20013d}. Template models allow for longer planning horizons and reduce computational costs, and in many ways have enabled the expanded use of legged robots in recent years. 

Despite the widespread usage of template models, however, surprisingly little is known about formal connections between templates and full-order rigid-body robot models. This means that whole-body controllers used to track template models include many heuristic design elements. While such methods can achieve good results in practice, they require extensive parameter tuning and are not always robust. 

Addressing the theoretical gap between template and whole-body models, as well as the performance issues this gives rise to in practice, is the focus of a growing body of literature. A strict approach was taken in \cite{poulakakis2009spring}, where the dynamics of a Spring-Loaded Inverted Pendulum (SLIP) were embedded as the the Hybrid Zero Dynamics \cite{westervelt2003hybrid} of an asymmetric hopper. Reachability analysis was used in \cite{liu2019leveraging} to constrain the template to enforce constraints in the full-order model. Our proposed approach is related most closely to \cite{kurtz2019formal}, which used the framework of approximate simulation to establish a formal relationship between the centroidal dynamics of a multi-link balancer and the LIP model. 

Approximate simulation, an extension of simulation/bi-simulation from formal methods to continuous systems, {is a relationship between two systems such that for some precision $\epsilon$, if the systems start $\epsilon$-close they can remain $\epsilon$-close \cite{girard2006hierarchical}}. In \cite{kurtz2019formal}, it was shown that establishing such a relation enables the projection of contact friction constraints to the template model, and the resulting controller enabled better recovery from push disturbances than a standard approach. 

\begin{figure}
    \begin{subfigure}{0.23\textwidth}
        \centering
        \includegraphics[width=0.9\linewidth]{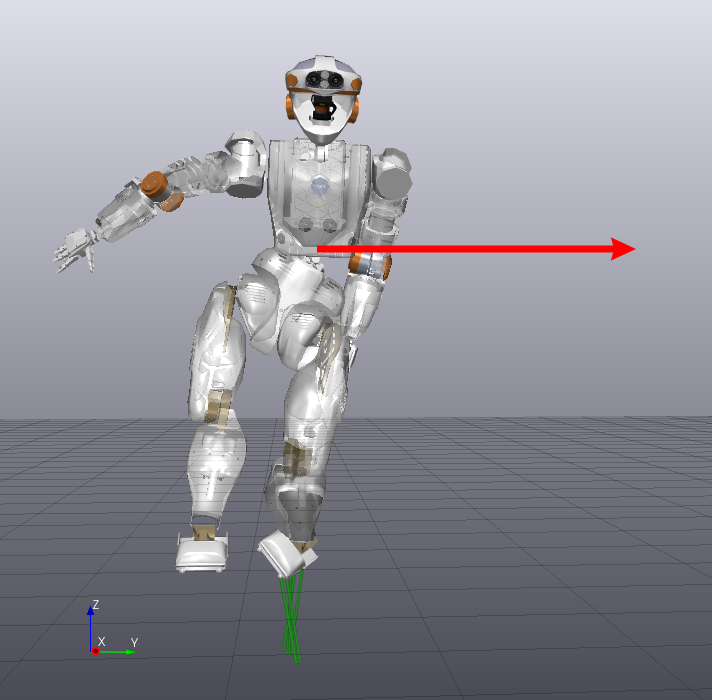}
        \caption{Recovering from pushes.}
        \label{fig:push}
        \vspace{5px}
    \end{subfigure}
    \begin{subfigure}{0.23\textwidth}
        \centering
        \includegraphics[width=0.9\linewidth]{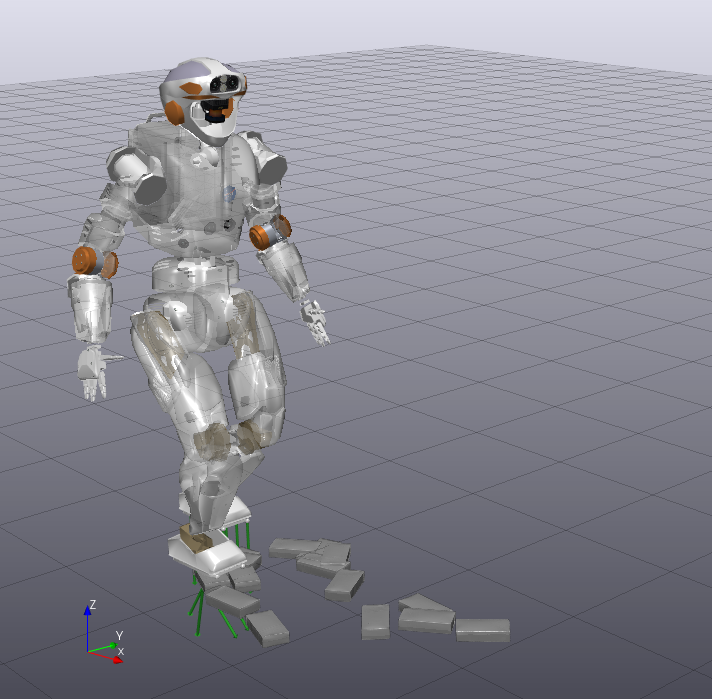}
        \caption{Walking over uneven terrain.}
        \label{fig:terrain}
        \vspace{5px}
    \end{subfigure}
    \begin{subfigure}{0.23\textwidth}
        \centering
        \includegraphics[width=0.9\linewidth]{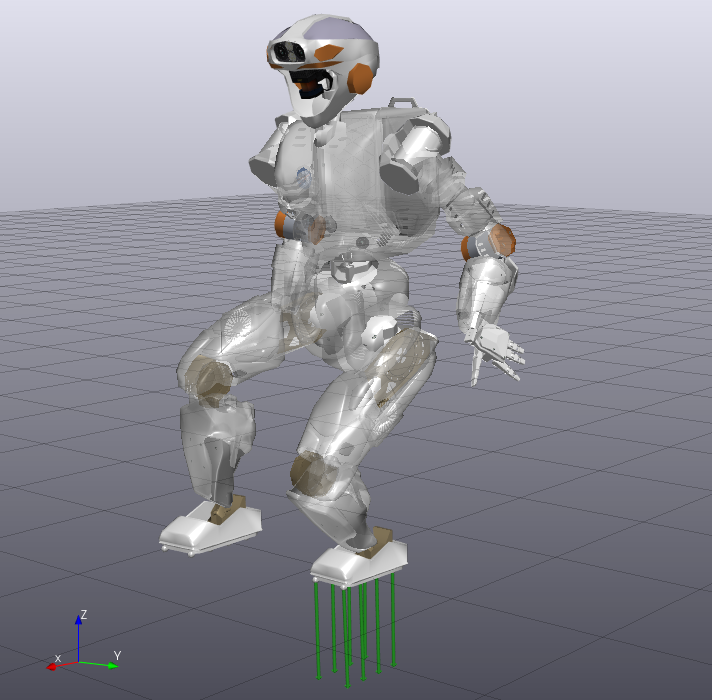}
        \caption{Robustness to modeling errors.}
        \label{fig:model}
    \end{subfigure}
    \begin{subfigure}{0.23\textwidth}
        \centering
        \includegraphics[width=0.9\linewidth]{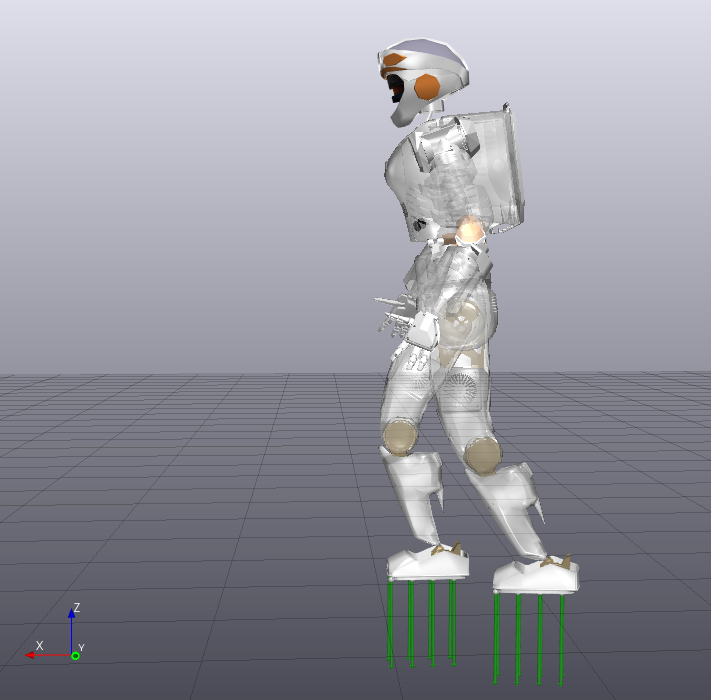}
        \caption{Walking with a high CoM.}
        \label{fig:high_com}
    \end{subfigure}
    \centering
    \caption{Simulation test scenarios considered in this letter. Our approach enforces approximate simulation and passivity, and consistently outperforms standard optimization-based whole-body control algorithms.}
    \label{fig:test_overview}
    \vspace{-1em}  
\end{figure}

These prior results have several important limitations, however. First, no formal connection was made between the CoM dynamics and the full rigid-body dynamics. Instead, it was assumed that desired CoM accelerations could be enforced via task-space feedback linearization: this enabled the use of existing results on approximate simulation for linear systems. Second, since full rigid-body dynamics were not considered explicitly, the approach of \cite{kurtz2019formal} does not account for subtasks essential to locomotion such as controlling swing foot trajectories. Finally, external disturbances arising from contact events and modeling errors were not considered. 

In this letter, we address each of these limitations. We establish approximate simulation between a simple linear CoM model and the (nonlinear) whole-body dynamics by drawing on the Hamiltonian structure of rigid-body dynamics. This enables a {hierarchical} approximate simulation relationship between the LIP model, CoM model, and whole-body model. We propose an optimization-based whole-body controller that enforces approximate simulation, manages complex whole-body constraints, and enables lower priority tasks when possible (e.g., swing foot tracking). We account for disturbances from impact events and modeling errors as in the recently proposed framework of robust approximate simulation \cite{kurtz2020robust}. In addition to enforcing approximate simulation, our proposed controller ensures passivity, suggesting superior safety and robustness properties \cite{folkertsma2017energy}.

These theoretical results are validated in simulation experiments with a 30 degree-of-freedom (DoF) Valkyrie humanoid. We find that our proposed whole-body controller is more robust to push disturbances, modeling errors, and uneven terrain than a standard whole-body controller. Furthermore, our approach continues to function effectively even in singular configurations, such as with straight legs, enabling more efficient walking with a high CoM. 

The remainder of this letter is organized as follows: background on approximate simulation is introduced in Section~\ref{sec:background}. Our main theoretical results are presented in Section~\ref{sec:main_results}, and simulation experiments are discussed in Section~\ref{sec:simulation}. We conclude with Section~\ref{sec:conclusion}.

\section{Background}\label{sec:background}

Approximate simulation is a relationship between two dynamical systems, $\Sigma_1$ and $\Sigma_2$:
\begin{equation}\label{eq:nonlinear_system}
    \Sigma_1 :
    \begin{cases}
        \xdot_1 = f_1(\x_1,\u_1,\mathbf{d}) \\
        \y_1 = g_1(\x_1)
    \end{cases},~~
    \Sigma_2 :
    \begin{cases}
        \xdot_2 = f_2(\x_2,\u_2) \\
        \y_2 = g_2(\x_2)
    \end{cases},
\end{equation}
where $\x_i \in \mathcal{X}_i \subseteq \R^{n_i}$ are the system states, $\u_i \in \mathcal{U}_i \subseteq \R^{p_i}$ are the control inputs, $\mathbf{d} \in \R^d$ is a disturbance signal representing external disturbances or modeling errors, and $\y_i \in \R^m$ are the system outputs. Note that the states may be of different sizes but the outputs must be of the same size. We typically consider $\Sigma_1$ to be a more complex system model (e.g., full rigid-body dynamics) and $\Sigma_2$ to be a simpler model (e.g., template dynamics). {\em $\Sigma_1$ approximately simulates $\Sigma_2$} if {there is some $\epsilon>0$ such that $\Sigma_1$ can always remain $\epsilon$-close to $\Sigma_2$}:
{
\begin{definition}[Approximate Simulation \cite{girard2006hierarchical}]
    A relation $\mathcal{R} \subseteq \mathcal{X}_1 \times \mathcal{X}_2$ is an approximate simulation relation of precision $\epsilon$ between $\Sigma_1$ and $\Sigma_2$ if for all $(\x_1^0,\x_2^0) \in \mathcal{R}$,
    \begin{enumerate}
        \item $\| g_1(\x_1^0) - g_2(\x_2^0)\| \leq \epsilon$
        \item For all state trajectories $\x_2(\cdot)$ of $\Sigma_2$ such that $\x_2(0) = \x_2^0$, there exists a state trajectory $\x_1(\cdot)$ of $\Sigma_1$ such that $\x_1(0) = \x_1^0$ and satisfying $(\x_1(t),\x_2(t)) \in \mathcal{R} ~\forall t \geq 0$.
    \end{enumerate}
    The relation $\mathcal{R}$ is a \textit{robust} approximate simulation relation if the above properties hold for any disturbance $\mathbf{d}$ in some set $\mathcal{D} \subseteq \R^d$.
\end{definition}
}

Typically, a controller is designed for the simplified model $\Sigma_2$: the full-order model $\Sigma_1$ can then also be guaranteed to complete the given task {with $\epsilon$ precision.}

{The most common way of certifying (robust) approximate simulation is by finding a Lyapunov-like (robust) simulation function \cite{girard2011approximate,kurtz2020robust} which certifies State-Independent Input to Output Stability (SIIOS) \cite{sontag2000lyapunov} of the joint system. This is a sufficient but not necessary condition for approximate simulation: SIIOS induces a uniform ultimate bound \cite{khalil2002nonlinear} on the output error that is independent of the initial condition and goes to zero in the absence of disturbances and exogenous inputs (i.e., $\mathbf{d} = 0, \mathbf{u}_2 = 0$), but these stronger stability conditions are not necessary for approximate simulation.}

{Establishing approximate simulation can be difficult for nonlinear systems}. For linear systems, there are well-established conditions for the existence of quadratic simulation functions \cite{girard2009hierarchical,kurtz2020robust}. Systematic design procedures for nonlinear systems remain an area of ongoing research, though state-of-the-art techniques based on Sum-of-Squares programming \cite{smith2019continuous} do not currently scale to high dimensional systems like legged robots. To address this challenge, we draw inspiration from energy shaping {and passivity-based control \cite{folkertsma2017energy,henze2016passivity}.}

{Specifically, we certify robust approximate simulation by finding a Lyapunov-like function $\V(\x_1,\x_2)$ that bounds the output error, and an interface $\u_1 = u_\V(\x_1,\x_2,\u_2)$ that specifies inputs to $\Sigma_1$ such that $\y_1$ tracks $\y_2$. Our proposed $\V$ does not certify SIIOS; nonetheless, we are able to establish an approximate simulation relation $\mathcal{R}$ as a sub-level set of $\mathcal{V}$. This development is an important first step toward a longer-term goal of template-based planning with full-model guarantees. The resulting controller guided by this theory is also found to have passivity properties that lead to improved practical effectiveness over state-of-the-art QP-based control methods.}

\section{Main Results}\label{sec:main_results}

\subsection{Model Hierarchy}

\begin{figure}
    \centering
    \begin{tikzpicture}[thick]
        \node[draw, minimum width=4.5cm] (template) at (0,0) {Template (LIP) Model, $\Sigma_3$};
        \node[draw, below=0.5 of template, minimum width=4.5cm] (task) {CoM Model, $\Sigma_2$};
        \node[draw, below=0.5 of task, minimum width=4.5cm] (anchor) {Whole-Body Model, $\Sigma_1$};
        
        \draw[<-] (template) -- node[right] {\textit{linear interface (\ref{eq:lip_com_interface})}} (task);
        \draw[<-] (task) -- node[right] {\textit{energy shaping interface (\ref{eq:interface})}} (anchor);
    \end{tikzpicture}
    \caption{Model hierarchy considered in this approach. Our primary contribution is in establishing a robust approximate simulation relation between the intermediate CoM model and the whole-body model. }
    \label{fig:model_hierarchy}
    \vspace{-1.5em}
\end{figure}
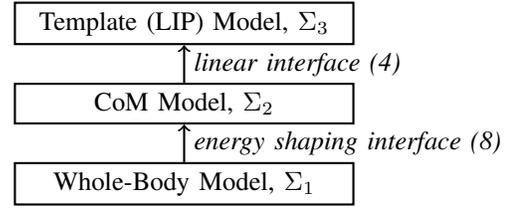

In our proposed control approach, we use the model hierarchy shown in Figure \ref{fig:model_hierarchy}. The whole-body model $\Sigma_1$ approximately simulates an intermediate CoM model $\Sigma_2$, which in turn approximately simulates the LIP model $\Sigma_3$. 

At the highest level of abstraction, the LIP model $\Sigma_3$ is used to generate a constant-height CoM trajectory consistent with a set of pre-planned footsteps. The LIP dynamics are given by \cite{kajita20013d}
\begin{equation}
    \begin{gathered}\label{eq:lip_dynamics}
        \xdot_3 = \mathbf{A}_{lip}\x_3 + \mathbf{B}_{lip}\u_3, ~~~~ 
        \y_3 = \mathbf{C}_{lip}\x_3,
    \end{gathered}
\end{equation}
where $\x_3 = \begin{bmatrix}(\mathbf{p}_{com}^{lip})^T & (\mathbf{v}_{com}^{lip})^T\end{bmatrix}^T \in \R^6$ is the position and velocity of the CoM, $\u_3 \in \R^2$ is the CoP, and $\y_3 = \mathbf{p}_{com}^{lip}$.

At the intermediate level of abstraction, we consider a simple single-integrator model of the CoM,
\begin{equation}\label{eq:com_dynamics}
        \xdot_2 = \u_2, ~~~~
        \y_2 = \x_2,
\end{equation}
where $\x_2 \in \R^3$ represents the CoM position. This model adds flexibility over the LIP (\ref{eq:lip_dynamics}) in the sense that the CoM is not constrained to a constant height. Noting that 
(\ref{eq:lip_dynamics}) and (\ref{eq:com_dynamics}) are linear systems, the interface
\begin{equation}\label{eq:lip_com_interface}
    \u_2 = \mathbf{v}_{com}^{lip} + \K(\x_2 - \mathbf{p}_{com}^{lip})
\end{equation}
enforces a robust approximate simulation relation between systems (\ref{eq:lip_dynamics}) and (\ref{eq:com_dynamics}), where $\K$ is any Hurwitz matrix \cite{kurtz2020robust}.

At the lowest level of abstraction, we consider the full multi-contact rigid-body dynamics of an $n$-link robot
\begin{equation}\label{eq:rigid_body_dynamics}
    \M(\q)\qddot + \mathbf{C}(\q,\qdot)\qdot + \btau_g = \mathbf{S}^T\btau + \sum\J_{c_j}^T\mathbf{f}_{c_j} + \btau_{ext},
\end{equation}
where $\q \in \R^{n+6}$ are generalized coordinates and $\btau$ are controlled joint torques. $\M(\q)$ is the positive definite mass matrix, $\mathbf{C}(\q,\qdot)\qdot$ collects Coriolis and centripetal terms, $\btau_g$ is the torque due to gravity, $\mathbf{f}_{c_j} \in \R^3$ are forces at contact points $c_j$, $\J_{c_j}$ are contact Jacobians, and $\btau_{ext}$ are external (disturbance) torques. {The robot's total mass is $m$.}

These whole-body dynamics form a nonlinear system with 
\begin{equation}\label{eq:wholebody_dynamics}
    \x_1 = 
    \begin{bmatrix}
        \q^T &
        \qdot^T
    \end{bmatrix}^T, ~~~~~
    \y_1 = \x_{com}(\q),
\end{equation}
where $\x_{com}$ is the position of the CoM. We denote the CoM Jacobian as $\J$ (i.e., $\xdot_{com} = \J\qdot$). 
{
\begin{remark}
    Some information is lost between the LIP model ($\Sigma_1$) and the CoM model ($\Sigma_2$), as the LIP includes accelerations but the CoM does not. While this leads to some degradation in tracking performance, it also enables potential energy shaping methods \cite{folkertsma2017energy} to certify approximate simulation. It also ensures that the resulting whole-body controller is numerically well-conditioned even in singular configurations, such as when the robot's legs are fully extended. This would not be the case if CoM accelerations were tracked directly using an inverse dynamics approach. 
\end{remark}
}

\subsection{Energy Shaping for Approximate Simulation}\label{subsec:energy_shaping_as}

As discussed in Section \ref{sec:background}, establishing approximate simulation for nonlinear systems is a difficult and open problem. In this section, we show that the Hamiltonian structure of $\Sigma_1$ (\ref{eq:wholebody_dynamics}) can be used to certify that $\Sigma_1$ approximately simulates the intermediate CoM model $\Sigma_2$ (\ref{eq:com_dynamics}). 

We first introduce the control transformation 
\begin{equation}
    \u_1 = \mathbf{S}^T\btau + \sum\J_{c_j}^T\mathbf{f}_{c_j}.
\end{equation}
The resolution of particular joint torques $\btau$ and contact forces $\mathbf{f}_{c_j}$ will be discussed in Section \ref{subsec:wholebody_control}. With this control transformation, energy shaping principles can be used to design a {Lyapunov-like function $\V$} and an interface $u_\V$ that certify robust approximate simulation as follows:

\begin{theorem}\label{theorem:robust_as}
    Assume that $\x_1$ and $\x_2$ are constrained to compact sets $\mathcal{X}_1$, $\mathcal{X}_2$. Let $\kappa > 0$ be a scalar constant, and $\K_D \succ 0$ be a symmetric matrix. Then the interface
    \begin{equation}\label{eq:interface}
        \u_1 = \btau_g - 2 \kappa \J^T (\y_1 - \y_2) - \K_D(\qdot - \J^T\u_2) 
    \end{equation}
    certifies that $\Sigma_1$ (\ref{eq:wholebody_dynamics}) robustly approximately simulates $\Sigma_2$ (\ref{eq:com_dynamics}) {via the Lyapunov-like function}
    \begin{equation}\label{eq:simulation_fcn}
        \V(\x_1,\x_2) = \begin{bmatrix} \qdot \\ \y_1 - \y_2 \end{bmatrix}^T
        \begin{bmatrix} \frac{1}{2\kappa}\M & \frac{\alpha}{2}\J^T\LL \\ 
                        \frac{\alpha}{2}\LL\J & \mathbf{I}
        \end{bmatrix}
        \begin{bmatrix} \qdot \\ \y_1 - \y_2 \end{bmatrix},
    \end{equation}
    where, $\LL = (\J\M^{-1}\J^T)^{-1} = m \mathbf{I}$ is the task-space inertia matrix, {and $\alpha$ is any constant chosen such that $0 < \alpha < \frac{1}{\sqrt{\kappa m/2}}$, and $\frac{1}{\kappa}\K_D - \alpha\J^T\LL\J \succeq \alpha \M$ uniformly over $\mathcal{X}_1$.}
\end{theorem}

This {Lyapunov-like} function essentially consists of the scaled closed-loop Hamiltonian under (\ref{eq:interface}) plus a cross term that ensures that $\Vdot < 0$ for large output errors $\|\y_1-\y_2\|$.

To prove Theorem \ref{theorem:robust_as}, we first need the following result regarding the structure of the rigid-body dynamics (\ref{eq:rigid_body_dynamics}):
\begin{lemma}[\cite{siciliano2010robotics}]\label{lemma:skew_symmetry}
    There exists a matrix $\mathbf{C}(\q,\qdot)$ such that (\ref{eq:rigid_body_dynamics}) holds and $\dot{\M}(\q,\qdot) - 2 \mathbf{C}(\q,\qdot)$ is skew-symmetric.
\end{lemma}

\begin{proof}[Proof of Theorem \ref{theorem:robust_as}]
    We will establish {a robust approximate simulation relation $\mathcal{R}$ as a sub-level set of $\V$}.
    
    First, we show that $\V \geq 0$. The Schur complement of 
    $\mathbf{P} = \begin{bmatrix} 
        \frac{1}{2\kappa}\M & \frac{\alpha}{2}\J^T\LL \\ 
        \frac{\alpha}{2}\LL\J & \mathbf{I}
    \end{bmatrix}$ is given by
    $\mathbf{I} - \frac{1}{2}\kappa\alpha^2\LL$.
    From this, the condition $\alpha < \frac{1}{\sqrt{ \kappa m / 2}}$ ensures $\V \geq 0$. Furthermore, we can see that $\mathbf{P}(\q) \succeq \beta \mathbf{I}$, where $\beta>0$ is the minimum eigenvalue of $\mathbf{P}(\q)$ taken over $\mathcal{X}_1$. This ensures $\beta\|\y_1-\y_2\|^2 \leq \V$, and so $\V$ is positive and bounds the output error.  
   
    Note that $\V$ can be written as 
    \begin{equation*}
        \V = \frac{1}{2\kappa}\qdot^T\M(\q)\qdot + U^{des}(\x_1,\x_2) + \alpha\qdot^T\J^T\LL(\y_1-\y_2),
    \end{equation*}
    where the first two terms consist of a kinetic energy term $\frac{1}{2\kappa}\qdot^T\M(\q)\qdot$, and a desired potential $U^{des}(\x_1,\x_2) = \|\y_1 - \y_2\|^2$. These two terms are a re-scaling of the closed-loop Hamiltonian
    \begin{equation}\label{eq:hamiltonian}
       H = \frac{1}{2}\qdot^T\M(\q)\qdot + \kappa\|\y_1 - \y_2\|^2.
    \end{equation}
    
    The Hamiltonian term evolves as follows:
    \begin{align*}
        \frac{1}{\kappa}\dot{H} =& \frac{1}{2\kappa}\qdot^T\dot{\M}(\q)\qdot + \frac{1}{\kappa}\qdot^T\M(\q)\qddot + \dot{U}^{des} \\
        =&\frac{1}{2\kappa}\qdot^T\dot{\M}\qdot + \frac{1}{\kappa}\qdot^T(\u_1 + \btau_{ext} - \mathbf{C}\qdot - \btau_g) + \dot{U}^{des} \\
        =&\frac{1}{\kappa}\qdot^T(\u_1 + \btau_{ext} - \btau_g) + \frac{\partial U^{des}}{\partial \q}\qdot +  \frac{\partial U^{des}}{\partial \x_2}\u_2
    \end{align*}
    by Lemma \ref{lemma:skew_symmetry}. Applying the interface (\ref{eq:interface}) and noting that $\left[\frac{\partial U^{des}}{\partial \q}\right]^T = 2\J^T(\y_1 - \y_2)$, we have that
    \begin{align}\label{eq:scaled_hdot}
        \frac{1}{\kappa}\dot{H} &= \frac{1}{\kappa}[\J\K_D\qdot]^T\u_2 + \frac{1}{\kappa}\qdot^T\btau_{ext} + \frac{\partial U^{des}}{\partial \x_2}\u_2 - \frac{1}{\kappa}\qdot^T\K_D\qdot.
    \end{align}
    
    {With the inclusion of the cross-term $\alpha\dot{\q}^T\J^T\LL(\y_1-\y_2)$},
    {
    \begin{multline*}
        \Vdot = -\qdot^T(\frac{1}{\kappa}\K_D - \alpha\J^T\LL\J )\qdot -2\alpha\kappa\|\y_1-\y_2\|^2 \\
        - 2\alpha^2\kappa(\y_1-\y_2)^T\LL\J\qdot + (OT),
    \end{multline*}
    where the ``other terms'' ($OT$) are given by
    \begin{equation*}
        OT = \z_1^T(\y_1-\y_2) + \z_2^T \u_2 + \frac{1}{\kappa}\qdot^T {\btau}_{ext},
    \end{equation*}
    \begin{align*}
        \z_1^T &= -2 \u_2^T + \alpha\qdot^T(\dot{\J}^T\LL) + \alpha(\u_2^T\J - \qdot^T)\K_D\M^{-1}\J^T\LL \\
        &~~~~~ + \alpha (\btau_{ext}^T - \qdot^T\mathbf{C}^T) \M^{-1} \J^T \LL + 2\alpha^2\kappa \qdot^T \J^T\LL \\
        \z_2^T &= \frac{1}{\kappa}\qdot^T\K_D\J^T - \alpha \qdot^T\J^T\LL .
    \end{align*}
    }
    { 
    Recalling that $\alpha \M \preceq \frac{1}{k}\K_D - \alpha\J^T\LL\J$, we have
    \begin{equation*}
        \dot{\V} \leq -2\alpha\kappa\V(\x_1,\x_2) + (OT).
    \end{equation*}
    Therefore if $\V \geq \frac{\|OT\|_\infty}{2\alpha\kappa}$, then $\dot{\V} \leq 0$, where $\|OT\|_\infty = \sup_{\x_1 \in \mathcal{X}_1, \x_2 \in \mathcal{X}_2} (OT)$. With this in mind, it is clear that
        \begin{equation*}
        \mathcal{R} = \left\{ \x_1, \x_2 \mid \V(\x_1,\x_2) \leq \frac{\|OT\|_\infty}{2\alpha\kappa} \right\}
    \end{equation*}
    is forward invariant. Furthermore, since $\|\y_1 - \y_2\|^2 \leq \frac{1}{\beta}\V$, $\mathcal{R}$ is an approximate simulation relation of precision 
    \begin{equation}\label{eq:precision}
        \epsilon = \sqrt{\frac{1}{\beta}\frac{\|OT\|_\infty}{2\alpha\kappa}}.
    \end{equation}
    }
\end{proof}    
{
\begin{remark}
    The proposed $\V$ does not establish SIIOS of the joint system as a standard simulation function would  \cite{girard2009hierarchical,smith2019continuous,girard2011approximate}. Furthermore, note that the precision $\epsilon$ depends on (a supremum over) the input to the reduced-order model, $\u_2$, external disturbances $\btau_{ext}$, and the state of the full-order model $\x_1$. This makes it difficult to compute $\epsilon$ a-priori. Nonetheless, we find in our simulation studies (see Section \ref{sec:simulation}) that the precision (\ref{eq:precision}), while conservative, is far from the edges of the whole state space. This suggests that it may be possible to establish SIIOS under the interface (\ref{eq:interface}), but we leave this as an open question.
\end{remark}
}

The basic idea behind this Theorem is to design an interface based on energy shaping control \cite{folkertsma2017energy} that drives $\y_1$ to $\y_2$. This interface (\ref{eq:interface}) is analogous to a PD controller with gravity compensation. The resulting closed-loop Hamiltonian provides a starting point for $\V$, which is derived by adding cross-terms that force $\Vdot \leq 0$ when $\V$ is large. 

\begin{remark}
    While a $\u_1$ that tracks a nominal CoM trajectory could also be chosen using task-space feedback linearization, our approach does not require a left inverse of the CoM Jacobian $\J$. This means that our interface (\ref{eq:interface}) is effective even when the robot is in a singular configuration (i.e., when the legs are fully extended). This enables walking with a higher CoM, as shown in Section \ref{sec:simulation}. 
\end{remark}

Beyond the fact that {the interface (\ref{eq:interface}) certifies a robust approximate simulation relation, it also enforces passivity: }

\begin{corollary}\label{cor:passivity}
    The interface (\ref{eq:interface}) ensures passivity of the closed-loop system (interconnection of $\Sigma_1$ and $\Sigma_2$) with respect to inputs $\u_2$ and external disturbances $\btau_{ext}$. 
\end{corollary}

\begin{proof}
    Consider the closed-loop Hamiltonian (\ref{eq:hamiltonian}). From (\ref{eq:scaled_hdot}), we have that
    \begin{equation}\label{eq:Hdot_inequality}
        \dot{H} \leq \left( 2\kappa(\y_1 - \y_2)^T + [\J\K_D\qdot]^T\right)\u_2 + \qdot^T\btau_{ext}.
    \end{equation}
    For $\u_2 = 0$, this certifies passivity of the mapping from external disturbances $\btau_{ext}$ to joint velocities $\qdot$. Similarly, in the absence of disturbances, (\ref{eq:Hdot_inequality}) certifies a passive map from the the input to the reduced-order model $\u_2$ to a unusual output $\left( 2\kappa(\y_1 - \y_2) + \J\K_D\qdot\right)$, composed of a scaled output error ($2\kappa(\y_1 - \y_2)$) and a scaled CoM velocity ($\J\K_D\qdot$).
\end{proof}

These passivity relationships essentially regulate how additional energy enters the system. This property has important implications for safety and robustness. Specifically, the feedback interconnection of passive systems is always passive, a property that does not hold for general notions of stability \cite{folkertsma2017energy}. Furthermore, passivity-based controllers tend to be remarkably robust to modeling errors and external disturbances { \cite{falugi2014model,albu2012energy,manjarekar2003robust,back2019robust},} a critical property for operation in uncertain real-world settings. {Unlike many passivity-based control schemes for humanoid robots \cite{henze2016passivity,mesesan2019dynamic,fahmi2019passive}, however, we obtain an explicit representation of the storage function which we can use to characterize the energy of the closed-loop system as well as to obtain formal guarantees as in Theorem~\ref{theorem:robust_as}.}

Combining Theorem \ref{theorem:robust_as} with existing results for linear systems \cite{girard2009hierarchical,kurtz2020robust}, we can obtain the following main result:

\begin{theorem}
    The whole-body model $\Sigma_1$ robustly approximately simulates the LIP model $\Sigma_3$.
\end{theorem}

\begin{proof}
    From Theorem \ref{theorem:robust_as}, $\Sigma_1$ robustly approximately simulates $\Sigma_2$. This means that for a fixed $\x_1(0)$ we can choose $\x_2(0)$ such that $\|\y_1(0)-\y_2(0)\| \leq \epsilon_1$. Then for any $\u_2(t)$, we have that $\|\y_1(t) - \y_2(t)\| \leq \epsilon_1$ for all $t \geq 0$ under the interface (\ref{eq:interface}). Following \cite{girard2009hierarchical}, we have that $\Sigma_2$ approximately simulates $\Sigma_3$. This approximate simulation relation is associated with a precision $\epsilon_2$. This means that for a fixed $\x_2(0)$, $\x_3(0)$ can always be chosen such that $\|\y_2(0)-\y_3(0)\| \leq \epsilon_2$. Then for any $\u_3(t)$, $\|\y_2(t) - \y_3(t)\| \leq \epsilon_2$ for all $t \geq 0$ under the interface (\ref{eq:lip_com_interface}). Thus for any $\u_3(t)$ there exists an $\epsilon>0$ and a whole-body control input $\u_1(t)$ such that:
    \begin{equation}
        \|\y_1-\y_3\| \leq \|\y_1-\y_2\| + \|\y_2-\y_3\| \leq \epsilon_1 + \epsilon_2 = \epsilon,
    \end{equation}
    and the Theorem holds.
\end{proof}

%
\subsection{Optimization-Based Whole-Body Control}\label{subsec:wholebody_control}

In the previous section, we showed how to select $\u_1$ to track the intermediate CoM model while ensuring approximate simulation and passivity. But choosing joint torques consistent with this $\u_1$ while meeting torque limits and contact friction constraints is non-trivial. To do so, we propose the following QP:
\begin{align}\label{eq:our_optimization}
    \min_{\substack{\u_2,\btau,\btau_0,\\ \qddot,\mathbf{f}_{ext}}} ~& w_1 \| \u_2 - \u_2^{des} \| + \\
    & ~~~ w_2\| \J_{foot}\qddot + \dot{\J}_{foot}\qdot - \ddot{\x}_{foot}^{des}\| +...\label{eq:foot_cost}\\
    \text{s.t. } & \M\qddot + \mathbf{C}\qdot + \btau_g(\q) = \mathbf{S}^T\btau + \sum \J^T_{c_j}\mathbf{f}_{c_j}\label{eq:dynamics_constraint} \\
                 & u_\V(\x_1,\x_2,\u_2) + \mathbf{N}^T\btau_0 = \mathbf{S}^T\btau + \sum \J^T_{c_j}\mathbf{f}_{c_j}\label{eq:interface_constraint} \\
                 & \J_{c_j}\qddot + \dot{\J}_{c_j}\qdot = 0\label{eq:accel_constraint} \\
                 & \mathbf{f}_{c_j} \in \text{ friction cones}\label{eq:friction_constraint} \\
                 & \underline{\btau} \leq \btau \leq \bar{\btau}\label{eq:torque_constraint}
\end{align}

The primary cost function (\ref{eq:our_optimization}) attempts to ensure that $\u_2$ matches a nominal input $\u_2^{des}$, given by the interface with the LIP model  (\ref{eq:lip_com_interface}). 
Additional weighted costs like (\ref{eq:foot_cost}) can be added to regulate lower-priority tasks like tracking a desired swing foot trajectory, minimizing angular momentum, and regulating torso orientation. 

The constraint (\ref{eq:dynamics_constraint}) enforces the robot's dynamics, and (for fixed $\q,\qdot$) is linear in the decision variables. (\ref{eq:interface_constraint}) ensures that the robot's CoM motion is determined by the interface (\ref{eq:interface}), which is linear in $\u_2$. We use the dynamically consistent null-space projector $\mathbf{N}$ \cite{khatib1987unified} to add some flexibility---all joint torques are not strictly determined by (\ref{eq:interface})---while ensuring that the CoM motion matches that requested by the interface. This {addition of $\mathbf{N}^T\btau_0$} enables lower-priority tasks like swing foot control. Contact points are fixed in place by (\ref{eq:accel_constraint}), while (\ref{eq:friction_constraint}) ensures that contact forces remain within their respective (Coulomb) friction cones. This friction constraint is linear if pyramidal inner approximations of the friction cones are taken. Finally, (\ref{eq:torque_constraint}) enforces torque limits. 

{
Note that since the CoM motion must match that requested by the interface (\ref{eq:interface}), Theorem~1 holds for any solution of this whole-body QP. At the same time, the use of $\u_2$ as a decision variable and the relaxation via the null-space projector $\mathbf{N}$ ensure that the left-hand-side of (\ref{eq:interface_constraint}) can take arbitrary values, and thus this equality constraint does not limit the feasibility of the QP. 
}

This proposed optimization scheme is closely related to the QPs widely used for whole-body control of high-DoF {torque-controlled} legged robots \cite{wieber2016modeling}. In such schemes, a desired CoM acceleration 
\begin{equation}\label{eq:fblin_law}
  \ddot{\x}_{com}^{des} = \dot{\mathbf{v}}_{com}^{lip} + \K_P^{lip}(\mathbf{v}_{com}^{lip}-\xdot_{com}) + \K_D^{lip}(\mathbf{p}_{com}^{lip}-\x_{com})
\end{equation}
that tracks the LIP is encoded as a cost, subject to similar dynamics, friction, contact, and torque constraints. This standard approach, however, does not enforce passivity or allow for any guarantees on tracking performance.  

As we will show in Section \ref{sec:simulation}, our approach is more robust to large push disturbances, uneven terrain, and modeling errors than these standard QPs. This is due to the fact that the interface constraint (\ref{eq:interface_constraint}) enforces passivity (see Corollary \ref{cor:passivity}), regulating the energy that can be injected into the system.

\section{Simulation Results}\label{sec:simulation}

We implemented the controller described above in simulation using a 30 DoF model of the NASA Valkyrie humanoid. Our implementation\footnote{Code: \url{https://github.com/vincekurtz/valkyrie_drake}} is in Python, using the Drake \cite{drake} simulation and dynamics platform. The whole-body QP (\ref{eq:our_optimization}) was solved at 200Hz with Gurobi \cite{gurobi}. 

{Solver time at each iteration was roughly 3ms (1.5ms for the standard approach), indicating that a suitably efficient implementation could run in real-time at roughly 300Hz. Furthermore, specialized active-set solvers with warm-starts have enabled real-time rates up to 5kHz for the standard approach \cite{kuindersma2014efficiently}. This suggests that such methods could achieve real-time control at roughly 2.5kHz under our approach. 
}

\begin{figure}
    \begin{subfigure}{0.48\linewidth}
        \centering
        \includegraphics[width=0.83\linewidth]{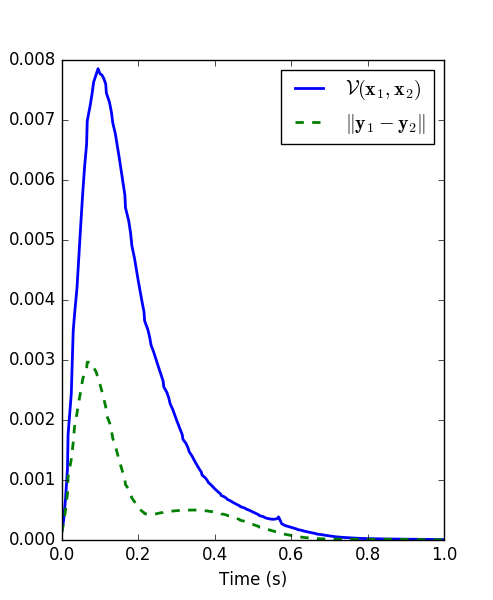}
        \caption{The {Lyapunov-like function $\V$} (\ref{eq:simulation_fcn}) decreases except when $\u_2$ is large, {keeping the output error $\|\y_1-\y_2\|$ bounded.}}
        \label{fig:standing_simulation_fcn}
    \end{subfigure}
    \begin{subfigure}{0.48\linewidth}
        \centering
        \includegraphics[width=0.87\linewidth]{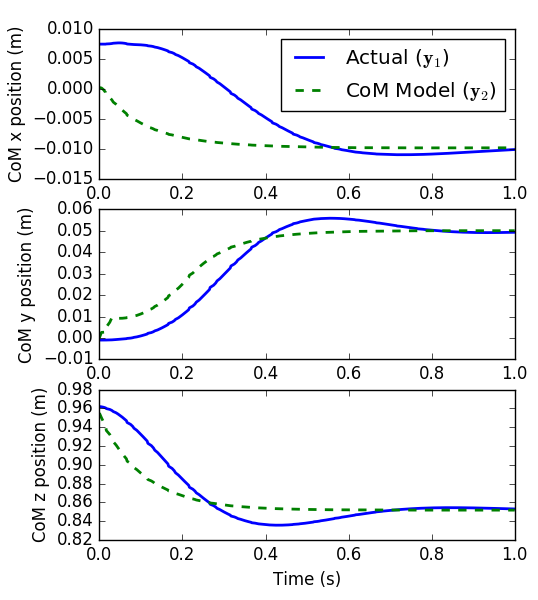}
        \caption{Both the linear CoM model and the true CoM converge to the setpoint.}
        \label{fig:standing_com_trajectories}
    \end{subfigure}
    \centering
    \caption{Using our approach to regulate the CoM while standing. }
    \label{fig:standing}
    \vspace{-1.5em}
\end{figure}

In Figure~\ref{fig:standing}, the robot moves its CoM to a desired position ($[-0.01~0.05~0.85]$m) while balancing in double support. First, a nominal trajectory of the LIP model $\x_3$ was generated, holding constant at the desired position. The CoM model $\x_2$ was initialized to match the true initial CoM position. The interface (\ref{eq:lip_com_interface}) was used to select a nominal input to the CoM model, $\u_2^{des}$. Then the whole-body QP (\ref{eq:our_optimization}) was solved to select joint torques. We used $\kappa=5000$ and $\K_D = 1000\mathbf{I}$. The robot's CoM moves smoothly to the desired position, while {$\V$} (\ref{eq:simulation_fcn}) decreases except when $\u_2$ is large. {The precision $\epsilon$ was estimated to be about $0.2m$.}
{
\begin{remark}
    We estimated $\epsilon$ by computing $\|OT\|_\infty$ as a maximum over executed trajectories (rather than the whole state space). More accurate data-driven estimates could be obtained by considering a range of behaviors performed offline using the proposed control strategy.
\end{remark}
}

We compare the performance of our approach with a standard whole-body QP {for torque-controlled robots}. For the standard QP, we generally followed the approach of \cite{wensing2013high}, using a LIP reference rather than a SLIP reference. We used a QP version of the controller presented in \cite{wensing2013generation} with the same weighted tasks as our approach. This method of tracking a template model with a QP-based whole-body controller is used broadly for {torque-controlled legged robots} \cite{wieber2016modeling}. 

Both methods used the same tuning parameters wherever applicable. For the standard approach, the parameters of (\ref{eq:fblin_law}) were tuned to match the step response of our approach. For each scenario, we specified a sequence of footsteps and swing foot trajectories a-priori. We then generated a corresponding LIP model trajectory \cite{kajita20013d}, and tracked it using either our approach or the standard whole-body QP. 

{
To illustrate the robustness of our method to state estimation error, we added Gaussian noise to the estimated floating base position, orientation, and velocity. We performed 6 trials with different noise levels, considering a trial successful if the robot walked for 5 seconds without falling. Results are shown in Table \ref{tab:state_estimation}. The first four columns use standard deviations for the position ($\sigma_p$), velocity ($\sigma_v$), and orientation ($\sigma_r$) noise based on \cite{flayols2017experimental}. Both our approaches were successful in all of these trials. The final two columns include high enough noise levels for each approach to fail. Our proposed approach withstood slightly more estimation noise than the standard approach. 
}
We then considered the four application scenarios shown in Figure \ref{fig:test_overview}: recovering from push disturbances, walking over uneven terrain, walking despite modeling errors, and walking with a high CoM. 

To test push recovery, we applied randomly generated pushes while the robot was walking. The time of each push was sampled uniformly from $[1.0, 3.5]$s. The push's magnitude was sampled uniformly from $[400,600]$N and applied for 0.05s in the $+y$ or $-y$ direction, each with 50\% probability. A trial was considered successful if the robot could walk for 5s without falling. The results are shown in {Table \ref{tab:main_trial_results}. Our approach was successful in 56/100 trials, while the standard method was successful only in 13 trials. The same push was applied for both approaches in each trial. Of the 87 trials in which the standard approach failed, our approach succeeded 47 times. On the other hand, of the 44 trials in which our approach failed, the standard approach succeeded only 4 times. }

\begin{table}[]
    \centering
    \begin{tabular}{|c||c|c|c|c|c|c|}
        \hline
        $\sigma_p$ (mm) & 3 & 3.7 & 26.0 & 34.9 & 40.0 & 50.0\\
        \hline
        $\sigma_r$ ($\degree$) & 0.5 & 0.5 & 0.9 & 0.9 & 1.5 & 1.5 \\
        \hline
        $\sigma_v$ (mm/s) & 14.2 & 18.5 & 82.2 & 107.8 & 200 & 250\\
        \hline
        \hline
        Ours & \cmark & \cmark & \cmark & \cmark & \cmark & \xmark\\
        \hline
        Standard & \cmark & \cmark & \cmark & \cmark & \xmark & \xmark\\
        \hline
    \end{tabular}
    \caption{{State estimation noise results (\xmark = fail, \cmark = success).}}
    \label{tab:state_estimation}
    \vspace{-1em}
\end{table}

\begin{table}[]
    \centering
    \begin{tabular}{|c||c|c|c|}
        \hline
        Task & Push Disturbances & Uneven Terrain & Modeling Error \\
        \hline
        Ours & 56 & 82 & 82 \\
        \hline
        Standard & 13 & 0 & 42 \\
        \hline 
    \end{tabular}
    \caption{{Successful simulation runs, each out of 100 trials}}
    \label{tab:main_trial_results}
    \vspace{-3em}
\end{table}

\begin{table}[]
    \vspace{-1em}
    \centering
    \begin{tabular}{|c||c|c|c|c|}
        \hline
        CoM Height (m)          & 0.95              & 1.00              & 1.02             & 1.05    \\
        \hline
        Ours  ($\times10^5$)    & $3.77$  & $2.88$  & $2.60$ & $2.18$ \\
        \hline
        Standard ($\times10^5$) & $3.51$  & $2.63$  & \xmark           & \xmark \\
        \hline
    \end{tabular}
    \caption{Approximate energy use ($\int\btau^T\btau dt$) for different CoM heights.}
    \label{tab:com_heights}
    \vspace{-1.5em}
\end{table}

\begin{figure}
    \centering
    \begin{subfigure}{0.48\linewidth}
        \centering
        \includegraphics[width=\linewidth]{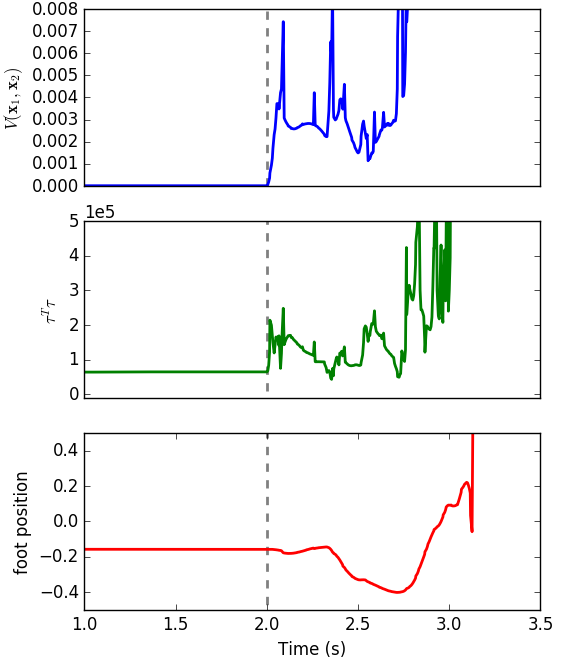} 
        \caption{Standard Approach. }
        \label{fig:push_standard}
    \end{subfigure}
    \begin{subfigure}{0.48\linewidth}
        \centering
        \includegraphics[width=\linewidth]{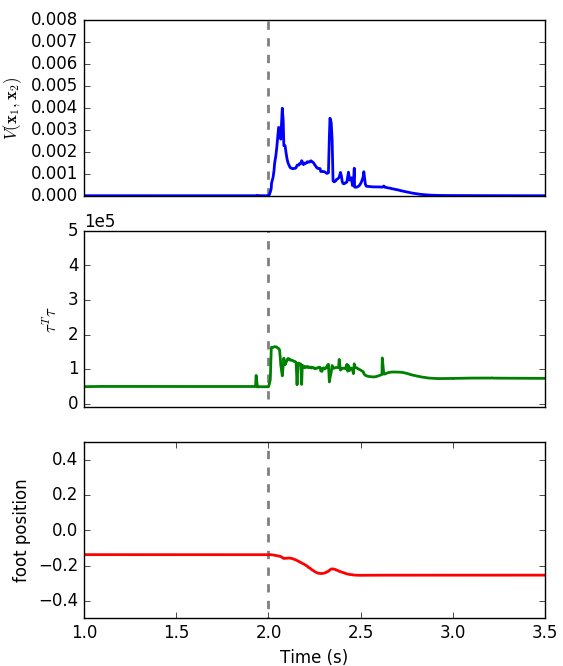} 
        \caption{Our Approach. }
        \label{fig:push_ours}
    \end{subfigure}
    \caption{{Storage function (top), sum of squared joint torques (middle), and right foot y-position (bottom) over time for a standing push recovery scenario. An 800N push disturbance was applied at $t=2$s. Our approach recovered from the push, while the standard approach did not.}}
    \label{fig:push_recovery}
    \vspace{-1.5em}
\end{figure}

{
A similar push recovery scenario, with an 800N push from standing, is illustrated in Figure~\ref{fig:push_recovery}. Our controller recovers from this push, while the standard controller does not. Figure~\ref{fig:push_recovery} includes plots of the storage function (\ref{eq:simulation_fcn}), which decays after the push under our approach and can be thought of as a measure of system energy, along with squared control inputs over time. Since the push results in a shifting of the stance foot, we also plot the y-position of the right foot over time. This illustrates how the inherent passivity of our approach allows the system to settle relatively quickly into a new equilibrium even after an unexpected shift in foot position. This ability to adapt to shifting footholds helps explain the effectiveness of our proposed approach in walking over rough terrain, since rough terrain induces frequent shifting of the stance feet. 
}

We tested robustness to uneven terrain by randomly placing 15 bricks in the robot's path. These bricks were fixed in place, but were smaller than the robot's foot, leading to frequent shifting of the stance foot while walking. A trial was considered successful if the robot could cross the bricks without falling. The results are shown in Table {\ref{tab:main_trial_results}}, with the same brick placement used for both approaches in each trial. {Our approach was successful in 82/100 trials}, while the standard approach was not successful in any of the trials. 

Empirically, the primary failure mode for our approach was when the swing foot got caught behind a brick. In contrast, the primary failure mode for the standard approach was when the stance foot shifted significantly, an event that occurred many times in each trial. The passivity of our approach helps explain our method's success in this regard: a shift in stance foot essentially injects energy into the system. Our controller's passivity ensures that the total system energy does not grow unbounded as a result {(see Figure~\ref{fig:push_recovery})}.

The third test scenario dealt with modeling error. In this scenario, the controller assumed the same robot model as in the trials above, but the true robot model was randomly perturbed. Specifically, the mass of each link, $m_i$, was modified according to
\begin{gather*}
    m_i^{new} = \max \left\{{\delta}, m_i + m_ir\right\}, ~~~
    r \sim \mathcal{N}(0,0.5),
\end{gather*}
{where a small $\delta > 0$ ensures that only positive link masses are generated.} A trial was considered successful if the robot could walk for 5s without falling. The same randomized model was used for both approaches in each trial. The results are shown in Table \ref{tab:main_trial_results}. {Our approach was successful in 82/100 trials, while the standard approach was successful in only 42. Of the 58 trials in which the standard approach failed, our approach succeeded 43 times. Of the 18 trials in which our approach failed, the standard approach succeeded 3 times.}

In the final test scenario, we considered walking with a higher CoM, which reduces torques at the knees, enabling more efficient locomotion \cite{griffin2018straight}. Standard whole-body control approaches do not perform well in configurations like that shown in Figure \ref{fig:high_com}, where the CoM is high enough that the back leg is fully extended. In such near-singular configurations, the numerical conditioning of whole-body QPs becomes poor, often resulting in extreme whole-body movements and a loss of balance. Our approach, however, regulates the system energy rather than trying to enforce specific accelerations, and is effective in these configurations. 

We recorded the integral of squared joint torques, a proxy for energy use, over 5s of walking for various CoM heights. Results are shown in Table \ref{tab:com_heights}. Our approach was successful with a higher CoM, enabling lower energy consumption. 

\begin{figure}
    \centering
    \includegraphics[width=0.8\linewidth]{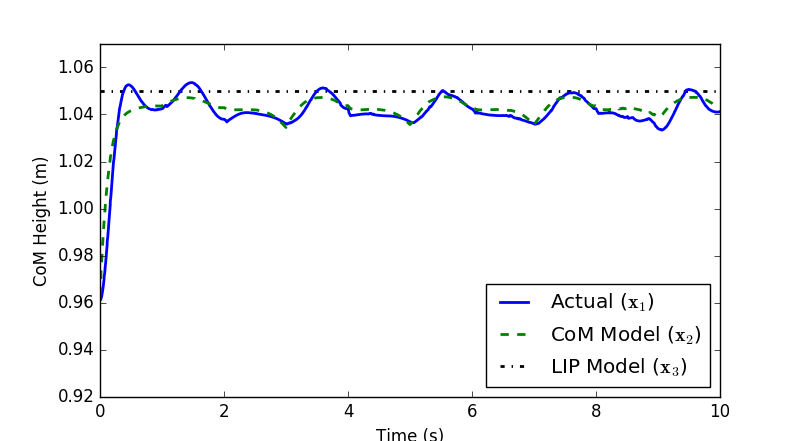}
    \caption{Output tracking for our approach walking with a high CoM. The energy-based nature of our approach enables effective operation in singular conditions, and approximate simulation relations between models allow for some deviation in the system outputs. Together, these properties enable more efficient walking with a high CoM.}
    \label{fig:com_trajectories}
    \vspace{-1em}
\end{figure}

High CoM walking is facilitated by the fact that approximate simulation allows for some deviation between the LIP CoM position ($\y_3$), the nominal CoM position ($\y_2$), and the true CoM position ($\y_1$). These three quantities are plotted in Figure \ref{fig:com_trajectories}. When the CoM is above the stance foot in single support, the desired LIP CoM height can be achieved ($t\approx\{3.5,5.5,7.5,9.5\}$s). In double support, however, the desired LIP CoM height is too high, and so the intermediate model ($\x_2$) height is adjusted downward ($t\approx\{3,5,7,9\}$s). This flexibility reduces the burden on template trajectories of being exactly replicated by the whole-body model, and is a natural byproduct of {\em approximate} simulation relationships.

\section{Conclusion}\label{sec:conclusion}

We introduced a template-based whole-body control strategy that enforces approximate simulation between the LIP model, a linear CoM model, and the whole-body dynamics. In addition to enabling formal guarantees of tracking performance, our approach is passive, more robust than standard methods to push disturbances, uneven terrain, and modeling errors, and allows for high CoM walking. Future work will focus on implementing these methods on a real robot{, deriving tighter estimates of the precision $\epsilon$,} and on extensions to nonlinear template models. In this last regard, we anticipate that the energy-based approaches put forward in this letter will continue to enable approximate simulation strategies to scale to more complex systems.

\bibliographystyle{IEEEtran}
\bibliography{references}

\end{document}